\documentclass[10pt,twocolumn,letterpaper]{article}

\usepackage[pagenumbers]{wacv} 

\usepackage{graphicx}
\usepackage{amsmath}
\usepackage{amssymb}
\usepackage{booktabs}
\usepackage{tabularray}
\UseTblrLibrary{diagbox}

\usepackage[pagebackref,breaklinks,colorlinks]{hyperref}

\usepackage{multirow,makecell}
\usepackage{colortbl}
\usepackage{pifont}
\usepackage{arydshln}
\usepackage{stfloats}
\usepackage[table]{xcolor}
\definecolor{1c}{rgb}{1,0,0} 
\definecolor{2c}{rgb}{0,0,1} 
\definecolor{3c}{rgb}{0.3,0.8,0.3} 

\usepackage[capitalize]{cleveref}
\crefname{section}{Sec.}{Secs.}
\Crefname{section}{Section}{Sections}
\Crefname{table}{Table}{Tables}
\crefname{table}{Tab.}{Tabs.}

\def\wacvPaperID{xxxx} 
\def\confName{WACV}
\def\confYear{2024}

\usepackage{amsmath,amsfonts,bm}

\def\eqref#1{equation~\ref{#1}}

\def\1{\bm{1}}

\def\vp{{\bm{p}}}

\def\vx{{\bm{x}}}

\def\vz{{\bm{z}}}

\def\mW{{\bm{W}}}

\DeclareMathAlphabet{\mathsfit}{\encodingdefault}{\sfdefault}{m}{sl}
\SetMathAlphabet{\mathsfit}{bold}{\encodingdefault}{\sfdefault}{bx}{n}

\def\sR{{\mathbb{R}}}

\def\emA{{A}}

\begin{document}

\title{Efficient Training for Visual Tracking with Deformable Transformer}

\author{Qingmao Wei, Bi Zeng, Guotian Zeng\\
Guangdong University of Technology\\
{\tt\small tsingmoe@gmail.com}
}
\maketitle

\begin{abstract}
  Recent Transformer-based visual tracking models have showcased superior performance. Nevertheless, prior works have been resource-intensive, requiring prolonged GPU training hours and incurring high GFLOPs during inference due to inefficient training methods and convolution-based target heads. This intensive resource use renders them unsuitable for real-world applications. In this paper, we present DETRack, a streamlined end-to-end visual object tracking framework. Our framework utilizes an efficient encoder-decoder structure where the deformable transformer decoder acting as a target head, achieves higher sparsity than traditional convolution heads, resulting in decreased GFLOPs. For training, we introduce a novel one-to-many label assignment and an auxiliary denoising technique, significantly accelerating model's convergence.  Comprehensive experiments affirm the effectiveness and efficiency of our proposed method.  For instance, DETRack achieves 72.9\% AO on challenging GOT-10k benchmarks using only 20\% of the training epochs required by the baseline, and runs with lower GFLOPs than all the transformer-based trackers.
\end{abstract}

\section{Introduction}
\begin{figure}[t]
  \centering
  \includegraphics[width=0.94\linewidth]{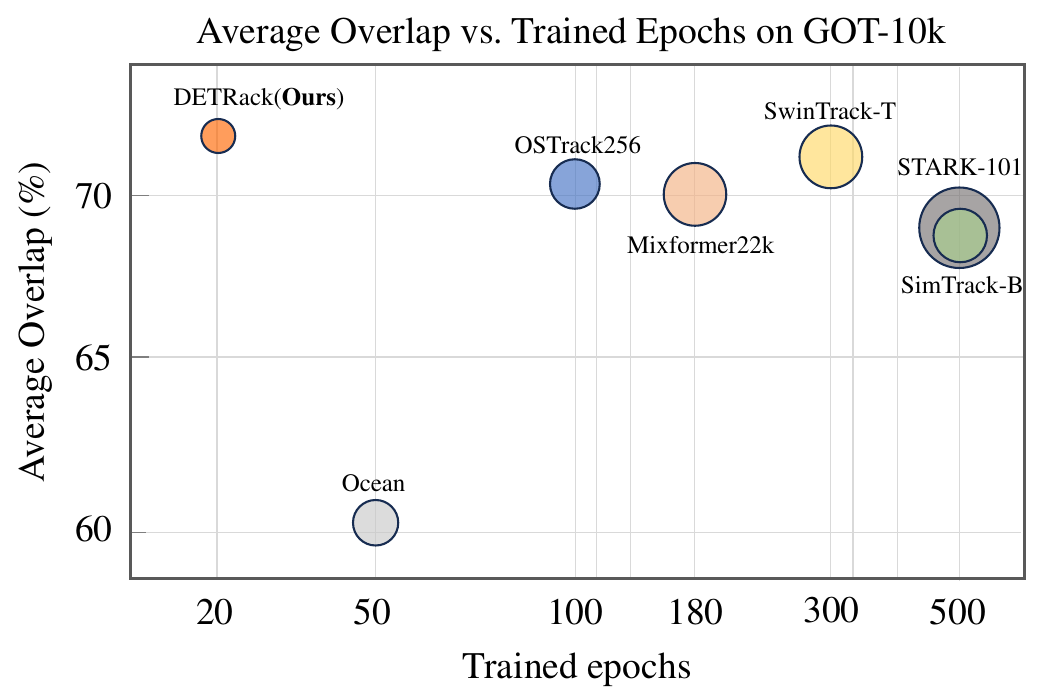}

   \caption{Comparison of our DETRack with other  trackers on GOT-10k benchmark in terms of trained epochs following the official one-shot protocol~\cite{GOT-10k}.  The bubble size represents the relative GFLOPs.  Our DETRack sets up a best trade-off among accuracy, trained epochs and running GFLOPs.  }
   \label{fig:title}
\end{figure}
\label{sec:intro}
Visual object tracking remains a critical challenge in computer vision, finding applications in diverse areas from surveillance to robotics and autonomous driving. The recent adoption of the Transformer~\cite{vanilla_selfattention} within visual trackers~\cite{transt,TMT,ostrack,mixformer,cswintt,simTrack,aiatrack} has introduced both innovation and complexity to this domain. Although deep learning advancements have elevated performance, training state-of-the-art (SOTA) trackers leveraging Transformers remains demanding, both in terms of time and computational resources. The significant GPU hours necessary for training a competitive tracker pose challenges, particularly for researchers with limited computational resources. Furthermore, the substantial parameters and GFLOPs associated with a SOTA tracker hinder their applicability in downstream tasks.


The present mainstream approaches for visual object tracking typically follow three primary stages: (i) deep neural network based feature extraction from the search and template images (ii) an integration module using either convolution or attention mechanisms for feature matching/fusion, and (iii)a head for bounding-box localization through customized heads for corner, center or scale estimation, and target classification. In some cases, the first two stages can be combined using a unified architecture, \eg Transformer Encoder and thus enjoying the powerful mask-image-modeling pretraining~\cite{mae,cae,simMIM}.  For accelerating running speed, sparsity is bringed into this unified process by some works.  Specifically, some image features, also called as tokens in the Transformer, can be dropped if they are considered irrelevant to the target to be tracked.  The sparicified process reduces about 66\% tokens compared to the original one~\cite{ostrack} before feeding the tokens into the head.  Those dropped tokens reduce about 20\% to 30\% GFLOPs in the encoder.  However, the prevailing head designs in the last stage are usually convolutional based, which require a fully 2D feature map as input.  Before feeding the sparicified tokens into the convolution head, those dropped ones must be padded (usually by zeros).  In the end, convolutional computaions are performed redundantly over padded dropped tokens.  On the other hand, the learning objective for classification in the prevailing head designs, such as center and corner require the model predicting a unique and sharp class map, which is difficult to optimize.  

\begin{figure}[t]
  \centering
  \includegraphics[width=0.94\linewidth]{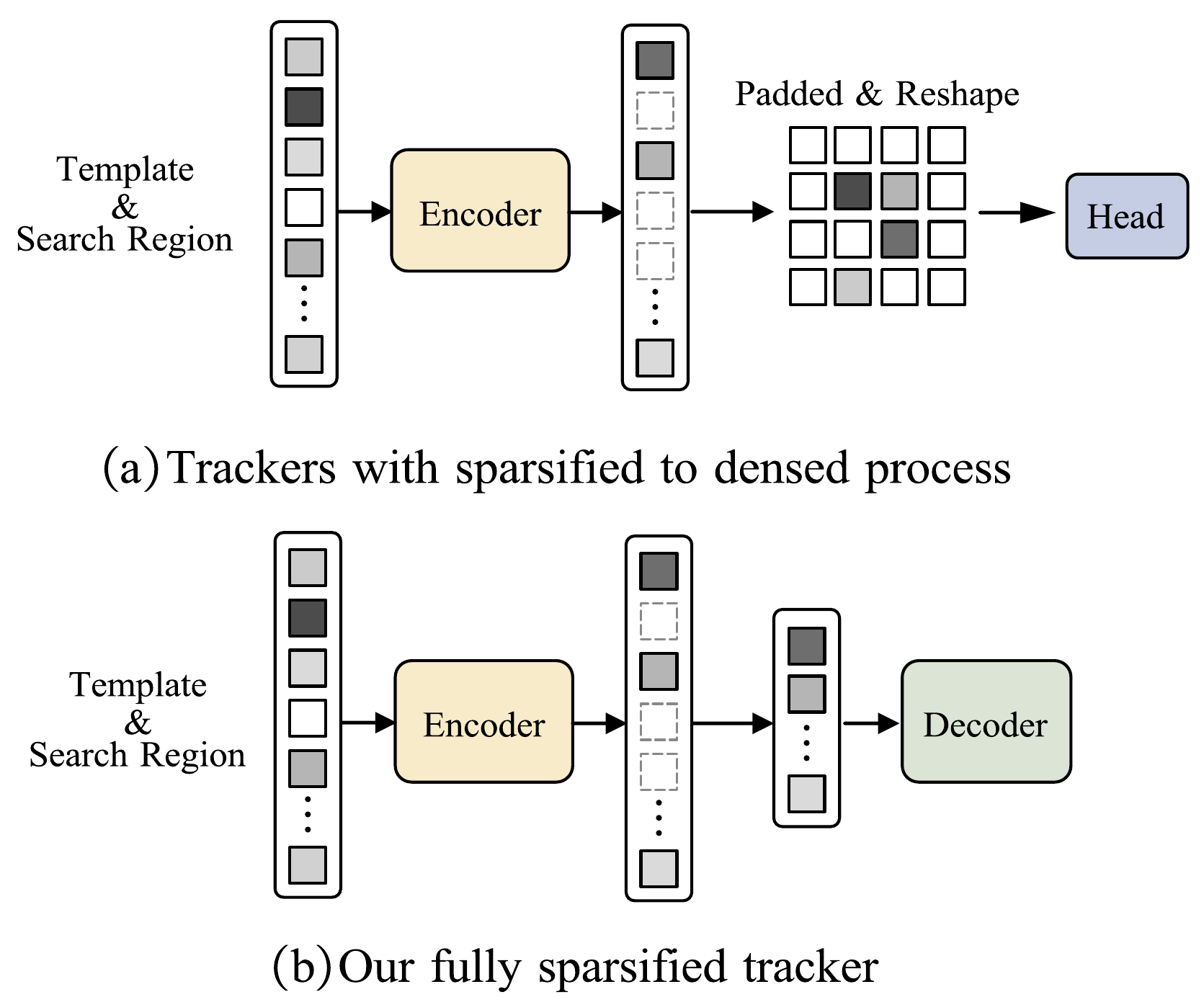}

   \caption{(a) Trackers with convolutional head have to padded the dropped features; (b) Our DETRack with the decoder maintains the full sparsity without redundant computation on the padded features. }
   \label{fig:sparsity}
\end{figure}

To tackle the redundant cumputation problem, we introduce the transformer decoder as the target head and thus make our model a encoder-decoder framework, dubbed as DETRack.   With capability of handling seqeunce of features, the decoder only deal with the reserved tokens after sparcified process by the encoder, avoiding redundant computation on the dropped tokens.  It enables our model keeping fully sparcified as shown in \cref{fig:sparsity}.  Insead of applying one-to-one Hungarian matching  during training like the other encoder-decoder designs in object detection~\cite{detr, dn-detr},  we found that a loose one-to-many label assignment significantly accelerating the training convergence.  We allow all the feature pixels within the GT bounding box to predict a positive classification score, which is expected to reflect the localization quality, \eg, IoU between the predicted bounding box and the GT.  For localization, we pick up the multiple predicted bounding box with the highest classification scores, thus introducing more supervision signal in the training.  Further more, we design a novel denoising branch as a auxiliary training strategy.  The noised GT bounding boxes as region proposals enrich the diversity of  training samples, which further acccelerate the training speed. 

Our extensive experiments validate the effectiveness and efficiency of our DETRack.  Specifically, compared to the most training-efficient transformer-based tracker OSTrack, our method further reduce the training epochs to 20\%, while achieving a higher performance with lower GFLOPs as shown in \cref{fig:title} and \cref{tab:main}.  Our main contributions are summarized as follows:
\begin{itemize}
  \item We propose a DETR-like encoder-decoder framework for visual object tracking without convolution head, thus maintain the computation efficiency for a sparsified backbone.
  \item We design a novel one-to-many label assignment during training, which significantly accelerate the trainging convergence.
  \item To avoid low quality prediction and further acccelerate training convergence, we introduce a denoising training strategy which bring in rich supervision signal.
\end{itemize}

\section{Related Work}

\subsection{Visual Tracking Paradims}
Over the past few years, Siamese trackers \cite{bertinetto2016fully,li2018high,atom} have gained much popularity. Typically, they adopt a two-stream pipeline to separately extract the features of the template and search region. Cross-relations between the two streams are modeled by additional correlation modules. To exploit the power of highly discriminative features, most Siamese trackers \cite{zhang2019deeper,li2019siamrpn++,bhat2019learning,Ocean} use a pre-trained deep neural networks as the backbone, \eg ResNet-50 \cite{he2016deep}, leave them frozen in training, thus they only need very few epochs to train the tracker.  Recently, the Transformer \cite{vaswani2017attention} architecture has achieved promising results in visual tracking and has become the de-facto choice for many high-performance trackers \cite{TMT,transt,STARK,TOMP,aiatrack,unicorn,SwinTrack}.

To further enhance the feature interaction, several attempts \cite{yu2020deformable,xie2022correlation,guo2022learning} have investigated cross-relation modeling inside the backbone. Recently, another thread of progress \cite{STARK,SwinTrack} concatenates the template and search tokens to conduct cross-relation modeling and self-relation modeling jointly. Inspired by these explorations, more recent trackers \cite{mixformer,ostrack,simTrack} adopt a one-stream pipeline to jointly extract the features and model the relations of both the template and search region by the self-attention mechanism. Based on this pipeline, they can utilize advanced pretrained models, \eg MAE \cite{mae}, instead of randomly initialized correlation modules for cross-relation modeling, thereby yielding a remarkable performance gain.


\subsection{DEtection with TRansformer (DETR)}
Carion \textit{et al.}~\cite{detr} proposed a Transformer-based end-to-end object detector named DETR (DEtection TRansformer) without using hand-designed components like anchor design and NMS. Many follow-up papers have attempted to address the slow training convergence issue of DETR introduced by decoder cross-attention. Deformable-DETR \cite{zhu2020deformable} proposed a deformable attention module to focus on important regions from multiple feature levels. DN-DETR~\cite{dn-detr} and DINO~\cite{dino} develop a denoising training strategy to helps the model to avoid duplicate outputs of the same target. Borrowing inspiration from DETR, STARK~\cite{STARK} casts target tracking as a bounding box prediction problem and solve it with an encoder-decoder transformer, in which the encoder models the global spatiotemporal feature dependencies between targets and search regions.  However, it still adopt a convolutional head after the transformer for the final prediction.  We also adopt the encoder-decoder framework for our tracker inspired by the DETR-like models, but totally get rid off the dense operation like convolution.  Besides, we apply a novel label assignment and denoising training strategy, significantly reduced the training GPU hours.

\section{Methods}

This section presents the DETRack method in detail. First, we briefly overview the model architecture of  our DETRack framework, including the encoder and the decoder. Then, we introduce the training strategy, including label assignment and denoising training.

\subsection{Preliminary}
DETR~\cite{detr} is an end-to-end Transformer-based framework for object detection.  In DETR, each query in the transformer is expected to associated with one object in the image.  
As studied in some variants of DETR~\cite{dab, dino}, it becomes clear that queries in DETR are formed by two parts: a positional part and a content part, which are referred to as positional queries and content queries in this paper. 
DAB-DETR~\cite{dab} explicitly formulates each positional query in DETR as a 4D anchor box $(x,y,w,h)$, where $x$ and $y$ are the center coordinates of the box and $w$ and $h$ correspond to its width and height. Such an explicit anchor box formulation makes it easy to dynamically refine anchor boxes layer by layer in the decoder.

DN-DETR~\cite{dn-detr} and DINO~\cite{dino} introduces a denoising (DN) training method to accelerate the training convergence of DETR-like models. It shows that the slow convergence problem in DETR is caused by the instability of bipartite matching. To mitigate this problem, DN-DETR proposes to additionally feed noised ground-truth (GT) labels and boxes into the Transformer decoder and train the model to reconstruct the ground-truth ones. 

\subsection{Model Architecture}
\begin{figure*}[t!]
  \centering
  \includegraphics[width=.95\textwidth]{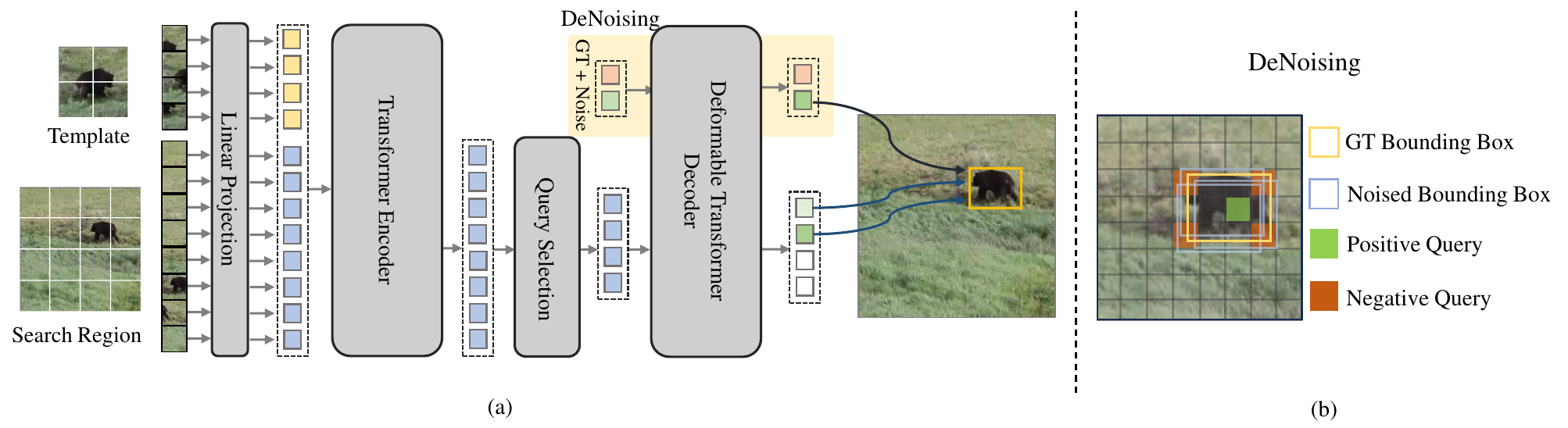}
  \caption{
    (a) Illustration of the architecture of our proposed DETRack. The denoising part in the decoder is only added during training.  (b) The label assignment for the denoising training.  The most center feature within the GT bounding box is picked as the positive object query in one denoising group, while the corner ones are selected as negative queries.
  }
  \label{fig:arch}
\end{figure*}
We use an encoder-decoder structure for learning and inference. Such a network architecture is widely used in modern visual recognition, especially DETR-like models~\cite{dn-detr,zhu2020deformable,dn-detr}.  Unlike DETR~\cite{detr} usually using a convolutional backbone for feature extraction in object detection, we simply use a single transformer encoder as backbone.

\textbf{Encoder.}
The encoder is identical to the one used in Vision Transformer (ViT).
We start by dividing a given image pair, comprising a template and a search region, into smaller image patches. Specifically, we denote the template image patch as $z \in \mathbb{R}^{3\times H_z \times W_z}$ and the search region patch as $x \in \mathbb{R}^{3\times H_x \times W_x}$.
These patches are then transformed into tokens, $\boldsymbol{H}_{z} \in \mathbb{R}^{D}$ and $\boldsymbol{H}_{x} \in \mathbb{R}^{D}$, using a linear projection layer. 
Then we add template and search tokens with positional embeddings, concatenate as $[ \boldsymbol{H}_{z} ; \boldsymbol{H}_{x}]$ and feed them into a plain ViT encoder to encode visual features.
For computation efficiency, some tokens can be dropped if they are considered irrelevant to the target in the encoder.  Here we simply adopt candidate elimination from OSTrack~\cite{ostrack} for the sparcified process.
 
\textbf{Query Selection.}
\label{sec:QS}
Tokens from the encoder directly serve as object queries in the decoder.
If there is no spacified process for dropping some tokens in the encoder, it will bring unacceptable computational and memory cost for the self-attention modules in the decoder.  To avoid this problem, we only select $K$ (we set $K=64$ in our implemention) tokens for the decoder no matter how many tokens are reserved after the encoder. Specically, each tokens output by the encoder directly predicts a bounding box and a foreground score by a linear layer.  The tokens with top-K score will be selected as content queires, and the bouding box of them are picked as region proposals.

\textbf{Decoder.}
There are mainly self attention and cross attention in each layer of the decoder.  The query elements for both types of attention modules are of object queries.  The self attention is a standard multi-head self attention, where object queries interact with each other.  In the cross attention modules, object queries interact with the features from the encoder, where the key elements are of the output tokens from the encoder.  The output object quiries by the decoder will be fed in to a MLP for the bounding box prediction and a linear layer for the foreground score prediction.  Unlike the encoder, we adopt the deformable attention~\cite{zhu2020deformable} in the decoder, which will be detailed describe in \cref{sec:decoder}.


\subsection{Deformable Transformer Decoder}
\label{sec:decoder}
\begin{figure}
  \centering
  \includegraphics[width=.8\linewidth]{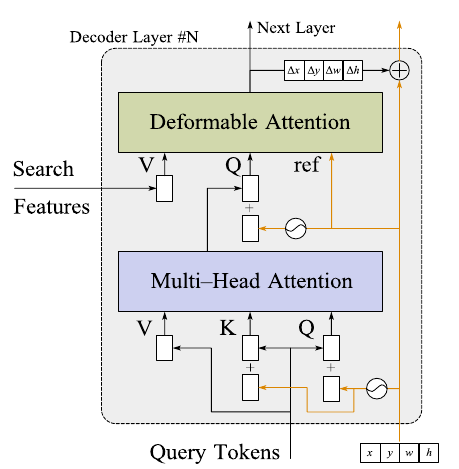}
  \caption{
    Illustration of one layer in the Deformable Transformer Decoder.  In each layer, the query tokens elementwise add the position embeddings generated by the box proposals to form the object queries for the attention module.  In the end of the layer, the predicted offset plus the box proposal as refined boxes are fed into the next layer. 
  }
  \label{fig:decoder}
\end{figure}

As shown in \cref{fig:decoder}, we formulate each object query in the deformable transformer decoder as two part: position part and content part in the decoder. The postion part are initialized from the region proposals (see \cref{sec:QS}) by $\sin$-$\cos$  embdeding in the fisrt layer.  For content part, we initialze it directly with the tokens from the encoder.  We simply element-wise add the two parts for the object queires.  After a standard multi-head attention, the core module in the decoder is deformable attention, in which the queires only attend to a small set of key sampling points around a reference point, regardless of the spatial size of the feature maps.  

\textbf{Deformable Attention.}
Given an input feature map $\vx \in \sR^{C \times H \times W}$, let $q$ index a query element with content feature $\vz_q$ and a 4-d reference box $(x,y,w,h)$, denoted as $\vp_q$. The deformable attention feature is calculated by
\begin{equation}
  \begin{aligned}
  &\text{DeformAttn}(\vz_q, \vp_q, \vx) =\\
  &\sum_{m=1}^{M} \mW_m \big[\sum_{k=1}^{K} \emA_{mqk} \cdot \mW'_m \vx(\vp_q + \Delta\vp_{mqk})\big],
  \end{aligned}
  \label{eq:single_deform_attn_fun}
  \end{equation}
where $m$ indexes the attention head, $k$ indexes the sampled keys, and $K$ is the total sampled key number ($K \ll HW$).
$\Delta\vp_{mqk}$ and $\emA_{mqk}$ denote the sampling offset and attention weight of the $k^\text{th}$ sampling point in the $m^\text{th}$ attention head, respectively.
The scalar attention weight $\emA_{mqk}$ lies in the range $[0, 1]$, normalized by $\sum_{k=1}^{K} \emA_{mqk} = 1$. $\Delta\vp_{mqk} \in \sR^4$ are of 4-d real numbers with unconstrained range. As $\vp_q + \Delta\vp_{mqk}$ is fractional, bilinear interpolation is applied in computing $\vx(\vp_q + \Delta\vp_{mqk})$. 
Both $\Delta\vp_{mqk}$ and $\emA_{mqk}$ are obtained via linear projection over the query feature $\vz_q$.

After a layer of decoder, the object queires are fed into a MLP to predict another 4-D offset $(\Delta x,\Delta y,\Delta w,\Delta h)$, as the adjustment to the target bounding box from the last layer's output.  As shown in \cref{fig:decoder}, the refined box output by the current layer serves as the 4-d reference box for the next layer of decoder. 

\subsection{Lable Assignment}
\label{assignment}
In object detection, DETR fundamentally addresses the issue of encouraging redundant predictions during training and the need for additional modules to eliminate such duplications during inference. DETR's one-to-one Hungarian matching promotes independent predictions during training. However, in single-object tracking, a given input image contains at most one target to be tracked. Thus, it's sufficient to merely select the prediction with the highest confidence from the final output.

Certain studies in object detection have indicated that encouraging repetitive predictions can significantly accelerate the convergence of training. We treat all tokens within the Ground Truth (GT) bounding box as potential positive samples.  However, simply assigning a hard label to the positive samples makes it difficult to select a high-quality prediction. To make the positive samples different, we introduce the localization quality (i.e., IoU score) into the classification score inpired by QFL~\cite{GFL}, 
where its supervision softens the standard one-hot category label and leads to a possible float target $y \in[0, 1]$.  Specifically, $y = 0$ denotes the negative samples with $0$ quality score, and $0 < y \leq 1$ stands for the positive samples with target IoU score $y$.  Therefore, the loss function for the classification is:
\begin{equation}
  \mathcal{L}_{\operatorname{cls}}(\sigma)=-|y-\sigma|^\beta((1-y) \log (1-\sigma)+y \log (\sigma)),
\end{equation}
where that $\sigma = y$ is the global minimum solution, standing for a accurate quality estimation. The parameter $\beta$ controls the down-weighting rate $|y-\sigma|^\beta$ smoothly.  In experiments, we set $\beta = 2$ following~\cite{GFL}.

For the box regression, to prevent supervision of low-quality tokens (e.g., tokens at the very edges or corners within the GT box), we filter out and retain only the top-k tokens based on predicted classification scores. We then compute the localization loss exclusively for the boxes output by these selected tokens.  We combine the $l_1$ loss and the generalized IoU loss~\cite{generalized_iou} as the training objective for the localization.  The loss function can be formulated as:
\begin{equation}
  \mathcal{L}_{\operatorname{loc}}(b)=\lambda_{\operatorname{G}} \mathcal{L}_{\operatorname{GIoU}}(b_i, \hat{b}_i ) + \lambda_{l_1} \mathcal{L}_{l_1}(b_i, \hat{b}_i ),
\end{equation}
where $b_i$ represents the groundtruth, and $\hat{b}_i$ represents the predicted box.  In experiments, we set the weights $\lambda_{\operatorname{G}} = 2$ and $\lambda_{l_1} = 5$.

\subsection{DeNoising Training}
\label{sec:denoising}
For each search image, we generate extra queires after the query selection by adding the noise to the GT.  The noised queires enrich the supervision signal for the decoder during the training process.  In implementation, the noised queires are also formulated as two part: position (box) and content (class).  

\textbf{Box Denoising.} We consider adding noise to boxes in two ways as in \cite{dn-detr}: center shifting and box scaling. We define $\lambda_1$ and $\lambda_2$ as the noise scale of these 2 noises. 1) \textbf{center shifting}: we add a random noise $(\Delta x, \Delta y)$, to the box center and make sure that $|\Delta x|<\frac{\lambda_1 w}{2}$, $|\Delta y|<\frac{\lambda_1 h}{2}$ , where $\lambda_1 \in (0, 1)$ so that the center of the noised box will still lie inside the original bounding box. 2) \textbf{box scaling}: we set a hyper-parameter $\lambda_2 \in (0, 1)$. The width and height of the box are randomly sampled respectively.  The noise scale for negative queires are set larger than that for the positive. 

For class denoising, the token positioned at the very center of the GT bounding box is chosen as the content part of the positive query, as illustrated in \cref{fig:arch}(b). To sidestep low-quality predictions, we select the corner tokens for the negative queries. We adopt a soft label assignment for alignment with the regular part in \cref{tab:assignment}. Specifically, the labels for positive queries are determined by the IoU between the predicted refined bounding box and the GT bounding box. In contrast, labels for negative queries are set to zero.

\textbf{Attention Mask.} It is proved to compromise the performance instead of improving it, without an attention mask preventing the GT information leaking from the denoising quires to un-denoising part~\cite{dn-detr}.  We devide the denoising queires as multiple groups, in which contains a positive and a negative query.  Queries in different groups and different parts cannot interact with each other.  The devision is performed by adding a mask to the self-attention module in the decoder as in ~\cite{dn-detr}.

\subsection{Training Objective}
The training objective for both denoising branch and regular branch are similar.  Specifically, we apply QFL in both regular branch and denoising branch for classification. For localization in denoising branch, we compute loss only on the positive queries.  As in DETR~\cite{detr}, we add auxiliary losses after each decoder layer and query selection module. Considering the loss of denoising branch after each decoder layer, our final loss can be written as generalized format:
\begin{equation}
  \mathcal{L}=\sum_{l=0}^{L_{\operatorname{dec}}}[\lambda_{\operatorname{cls}} (\mathcal{L}_{\operatorname{cls}} + \mathcal{L}_{\operatorname{cls}}^{\operatorname{DN}}) + \lambda_{\operatorname{loc}} (\mathcal{L}_{\operatorname{loc}}+\mathcal{L}_{\operatorname{cls}}^{\operatorname{DN}})],
\end{equation}
where $l=0$ denotes for the output before the query selection and $l=L_{\operatorname{dec}}$ represents for the final output of the decoder. We simply set the parameter $\lambda_{\operatorname{cls}}$ and $\lambda_{\operatorname{loc}}$ to 1 following~\cite{dino}.

\section{Experiments}

\begin{table*}[t!]
    \centering
    \resizebox{0.96\linewidth}{!}{
    \begin{tabular}{l|c|ccc;{1pt/1pt}c|ccc;{1pt/1pt}ccc;{1pt/1pt}c|c}
        \Xhline{2\arrayrulewidth}
        \multirow{2}{*}{Tracker}         & \multirow{2}{*}{Source} & \multicolumn{3}{c;{1pt/1pt}}{GOT-10k*} & \multicolumn{1}{l|}{\multirow{2}{*}{\#Epochs$^{**}$}} & \multicolumn{3}{c;{1pt/1pt}}{TrackingNet} & \multicolumn{3}{c;{1pt/1pt}}{LaSOT} & \multicolumn{1}{c|}{\multirow{2}{*}{\#Epochs$^{**}$}} & \multirow{2}{*}{GFLOPs}                                                                                                                            \\
                                         &                         & AO                                     & SR$_{0.5}$                                            & SR$_{0.75}$                               & \multicolumn{1}{l|}{}               & AUC                                                   & P$_\text{norm}$         & P                    & AUC                  & P$_\text{norm}$      & P                    & \multicolumn{1}{c|}{} &      \\
        \hline
        DETRack                          & Ours                    & \textcolor{1c}{72.9}                   & \textcolor{1c}{82.1}                                  & \textcolor{1c}{69.9}                      & \textbf{20}                         & \textcolor{1c}{83.2}                                  & \textcolor{1c}{88.3}    & \textcolor{1c}{83.1} & {69.0}               & \textcolor{2c}{78.9} & \textcolor{2c}{75.1} & \textbf{60}           & \textbf{15.6} \\
        OSTrack~\cite{ostrack}           & ECCV'22                 & \textcolor{3c}{71.0}                   & \textcolor{3c}{80.4}                                  & \textcolor{3c} {68.2}                     & 100                                 & \textcolor{2c}{83.1}                                  & 87.8                    & 82.0                 & \textcolor{3c}{69.1} & 78.7                 & \textcolor{1c}{75.2} & 300                   & 21.5 \\
        OSTrack\dag~\cite{ostrack}       & ECCV'22                 & \textcolor{2c}{71.5}                   & \textcolor{2c}{80.7}                                  & \textcolor{2c}{68.9}                      & 100                                 & \textcolor{3c}{83.0}                                  & 87.7                    & 82.0                 & 68.5                 & 78.1                 & \textcolor{3c}{74.9} & 300                   & 16.4 \\

        AiATrack~\cite{aiatrack}         & ECCV'22                 & 69.6                                   & 80.0                                                  & 63.2                                      & 100                                 & 82.7                                                  & 87.8                    & 80.4                 & 69.0                 & \textcolor{1c}{79.4} & 73.8                 & 300                   & 18.5     \\
        SimTrack~\cite{simTrack}         & ECCV'22                 & 68.6                                   & 78.9                                                  & 62.4                                      & 500                                 & 82.3                                                  & 86.5                    & -                    & \textcolor{1c}{69.3} & 78.5                 & 74.0                 & 500                   & 20.1     \\
        Unicorn~\cite{unicorn}           & ECCV'22                 & -                                      & -                                                     & -                                         & -                                   & \textcolor{3c}{83.0}                                  & 86.4                    & \textcolor{2c}{82.2} & 68.5                 & 76.6                 & 74.1                 & -                     & 33.5     \\
        MixFormer~\cite{mixformer}       & CVPR'22                 & 70.7                                   & 80.0                                                  & 67.8                                      & 180                                 & \textcolor{2c}{83.1}                                  & \textcolor{2c}{88.1}    & 81.6                 & \textcolor{2c}{69.2} & \textcolor{3c}{78.7} & 74.7                 & 550                   & 23.0     \\
        ToMP~\cite{TOMP}                 & CVPR'22                 & -                                      & -                                                     & -                                         & -                                   & 81.2                                                  & 86.2                    & 78.6                 & 67.6                 & 78.0                 & 72.2                 & 200                   &  -    \\
        CSWinTT~\cite{cswintt}           & CVPR'22                 & 69.4                                   & 78.9                                                  & 65.4                                      & 600                                 & 81.9                                                  & 86.7                    & 79.5                 & 66.2                 & 75.2                 & 70.9                 & 500                   &  19.3    \\
        STARK~\cite{STARK}               & ICCV'21                 & 68.0                                   & 77.7                                                  & 62.3                                      & 500                                 & 81.3                                                  & 86.1                    & 78.1                 & 66.4                 & 76.3                 & 71.2                 & 500                   &  18.5    \\
        KeepTrack~\cite{KeepTrack}       & ICCV'21                 & -                                      & -                                                     & -                                         & -                                   & -                                                     & -                       & -                    & 67.1                 & 77.2                 & 70.2                 & -                     &  -    \\
        AutoMatch~\cite{AutoMatch}       & ICCV'21                 & 65.2                                   & 76.6                                                  & 54.3                                      & -                                   & 76.0                                                  & -                       & 72.6                 & 58.3                 & -                    & 59.9                 & -                     &  -    \\
        TransT~\cite{transt}             & CVPR'21                 & 67.1                                   & 76.8                                                  & 60.9                                      & -                                   & 81.4                                                  & 86.7                    & 80.3                 & 64.9                 & 73.8                 & 69.0                 & -                     &  -    \\
        Alpha-Refine~\cite{Alpha-Refine} & CVPR'21                 & -                                      & -                                                     & -                                         & -                                   & 80.5                                                  & 85.6                    & 78.3                 & 65.3                 & 73.2                 & 68.0                 & -                     &  -    \\
        TMT~\cite{TMT}                   & CVPR'21                 & 67.1                                   & 77.7                                                  & 58.3                                      & -                                   & 78.4                                                  & 83.3                    & 73.1                 & 63.9                 & -                    & 61.4                 & -                     &  -    \\
        Ocean~\cite{Ocean}               & ECCV'20                 & 61.1                                   & 72.1                                                  & 47.3                                      & 50                                  & -                                                     & -                       & -                    & 56.0                 & 65.1                 & 56.6                 & 50                    &  -    \\
        DiMP~\cite{DiMP}                 & ICCV'19                 & 61.1                                   & 71.7                                                  & 49.2                                      & -                                   & 74.0                                                  & 80.1                    & 68.7                 & 56.9                 & 65.0                 & 56.7                 & -                     &  -    \\
        ATOM~\cite{atom}                 & CVPR'19                 & -                                      & -                                                     & -                                         & -                                   & 70.3                                                  & 77.1                    & 64.8                 & 51.5                 & 57.6                 & 50.5                 & -                     &  -    \\
        SiamRPN++~\cite{SiamRPN++}       & CVPR'19                 & 51.7                                   & 61.6                                                  & 32.5                                      & -                                   & 73.3                                                  & 80.0                    & 69.4                 & 49.6                 & 56.9                 & 49.1                 & -                     &  -    \\
        \Xhline{2\arrayrulewidth}
    \end{tabular}}
    \caption{State-of-the-art comparison on GOT-10k, TrackingNet and LaSOT. The best three results are shown in \textcolor{1c}{red} ,\textcolor{2c}{blue} and \textcolor{3c}{green} fonts, respectively. We use * to denote that the results on GOT-10k are obtained following the official one-shot protocol.  ** denotes the we calculate the effective number of epochs for $6\times10^4$ image-pairs sampled per epoch if the training detail is provided. \dag  ~indicates the tracker we replace the backbone with the same CAE~\cite{cae} pretrained ViT-Base model as our DETRack.}
    \label{tab:main}
\end{table*}
\begin{table*}[t]
    \footnotesize
    \centering
    \resizebox{\textwidth}{7.5mm}{
        \begin{tabular}{c|cccccccc|c}
            \Xhline{2\arrayrulewidth}
            Tracker                            & SiamRPN++\cite{SiamRPN++} & ATOM \cite{atom} & DiMP \cite{DiMP} & TransT \cite{transt} & TMT \cite{TMT} & STARK \cite{STARK} & ToMP \cite{TOMP} & OSTrack \cite{ostrack} & DETRack              \\
            \hline
            NFS \cite{kiani2017need}           & 50.2                            & 59.0                          & 62.0                         & \textcolor{3c}{65.7}              & \textcolor{blue}{66.5}         & 65.2                         & \textcolor{red}{66.9}             & 64.7                       & 65.3                 \\
            UAV123 \cite{mueller2016benchmark} & 61.3                            & 65.0                          & 65.4                         & \textcolor{1c}{69.1}              & 67.5                           & \textcolor{1c}{69.1}         & \textcolor{2c}{69.0}              & 68.3                       & \textcolor{3c}{68.7} \\

            \#Epochs                           & -                               & -                             & -                            & -                                 & -                              & 500                          & 200                               & 300                        & 60                   \\
            \Xhline{2\arrayrulewidth}
        \end{tabular}}
    \caption{Comparison with the state-of-the-art trackers on NFS and UAV123 in terms of AUC score.}

    \label{tab:addition}
\end{table*}
\begin{table}
    \centering
    \resizebox{0.95\linewidth}{!}{
        \begin{tabular}{c|c|l|l|c}
            \Xhline{2\arrayrulewidth}
            Sparsity & Head    & Params(M)                                                                    & FLOPs(G)                                                                     & AO(\%) \\
            \hline
            -        & conv    & 70.9                                                                         & 22.3                                                                         & 71.5     \\
            -        & decoder & \textbf{68.9} \fontsize{7.0pt}{\baselineskip}\selectfont{($\downarrow$ 2.0)} & \textbf{21.4} \fontsize{7.0pt}{\baselineskip}\selectfont{($\downarrow$ 0.9)} & 72.9     \\
            \hline
            CE       & conv    & 70.9                                                                         & 16.4                                                                         & 71.7     \\
            CE       & decoder & \textbf{68.9} \fontsize{7.0pt}{\baselineskip}\selectfont{($\downarrow$ 2.0)} & \textbf{15.6} \fontsize{7.0pt}{\baselineskip}\selectfont{($\downarrow$ 0.8)} & 72.9     \\
            \Xhline{2\arrayrulewidth}
        \end{tabular}}
    \caption{ Comparison of params, FLOPs, and performance(Average Overlap) on GOT-10k.
        The sparsified method CE indicates the Candidate Elimination~\cite{ostrack}.
    }
    \label{tab:flops}
    \vspace{-4mm}
\end{table}
\subsection{Implemention Details}
Our trackers are implemented using Python 3.8 and PyTorch 1.12. The models are trained on 2 RTX 3090 GPUs.  We test the model on a single RTX 2080Ti GPU.

\textbf{Model.}
We adopt the vanilla ViT-Base~\cite{vit} model and initialze it with CAE~\cite{cae} pre-trained weights on ImageNet ~\cite{imagenet_cvpr09} as the encoder of our DETRack.  We leave the weight of the decoder randomly initialized. The sizes of the template and search images are 128$\times$128 pixels and 256$\times$256 pixels, respectively. All the input images are split into 16$\times$16 patches. The hidden dimension in the encoder and the decoder are 768 and 256.  The dimension of tokens output by the encoder are transformed by a linear layer.  The bouding box and its offset in each decoder layer is obtain by three-layer perceptrons, which shares the parameter.  The classification score is predicted by a simple linear layer.    

\textbf{Training.}
\label{training}
For the GOT-10k~\cite{GOT-10k} benchmark, we only use the training split of GOT-10k following the one-shot protocols and train the model for 20 epochs.  
For the other benchmarks, the training splits of GOT-10k, COCO~\cite{COCO}, LaSOT~\cite{LaSOT} and TrackingNet~\cite{TrackingNet} are used for training in 60 epochs.
For video datasets, we sample the image pair from a random video sequence. For the image dataset COCO, we randomly select an image and apply data augmentations to generate an image pair. 
Common data augmentations such as scaling, translation, and jittering are applied on the image pair. 
The search region and the template are obtained by expanding the target box by a factor of 4 and 2, respectively.
The optimizer is the AdamW optimizer~\cite{adamw}, with the weight decay of 1e-4. 
The initial learning rate of the encoder and the decoder are 4e-5 and 4e-4, respectively.
We reduce the learning rate to 10\% in the last 20\% epochs.  
Each GPU holds 64 image-pairs, resulting a batch size of 128 in total.
Training takes $<$\textbf{2} hours for GOT-10k and $<$\textbf{6} hours for the other benchmarks on two RTX 3090 GPUs. Note that the training time of the previous most efficient transformer-based tracker, OSTrack, is about 8 hours, which is 4$\times$ that of our method.

\textbf{Testing.} 
During testing, we adopt Hanning window penalty to utilize positional prior like scale change and motion smoothness in tracking, following the common practice~\cite{SiamFC, SwinTrack, ostrack}. The output scores of each object queires by the decoder are reflected back on the 2D map and simply element-wise multiplied by the Hanning window with the same size, and we choose the box with the highest multiplied score as the target box.

\subsection{Comparison with State-of-the-art Trackers}
We compare our DETRack with state-of-the-art(SOTA) trackers on 3 different large-scale benchmarks and 2 small benchmarks, including GOT-10k, TrackingNet, LaSOT, UAV123 and NFS.  For fair comparison, the number of training epochs in \cref{tab:main} and \cref{tab:addition} of different trackers are counted with the same settings as in \cite{ostrack}, \eg, $6\times10^4$ image pairs sampled in each epoch.

\textbf{GOT-10k.}
GOT-10k~\cite{GOT-10k} is a large-scale dataset containing more than 10000 video segments of real-world moving objects. The object classes between train and test sets are zero-overlapped. We strictly follow the one-shot protocol to only train our model on the GOT-10k training split and evaluate the results through the evaluation server.  As presetented in \cref{tab:main}, DETRack improves all matrics by a large margin, \eg, 1.9\% in Average Overlap (AO) compared with OSTrack which indicates the capability in accurate discrimination and localization of unseen objects.  Notice that our tracker is trained with only 20 epochs, which is 20\% of the most training-efficient transformer-based traker OSTrack.

\textbf{TrackingNet.}
TrackingNet~\cite{TrackingNet} is a large-scale short-term tracking benchmark that provides more than 30000 video sequences with over 14 million boxes. The test split of TrackingNet contains 511 sequences without publicly available ground truth and covers diverse target classes andscenes. We submit the tracking results to the official evaluation server and make comparisons with previous SOTA
trackers in \cref{tab:main}.
The results show that our DETRack achieves 83.2\% in success score(AUC) and 83.1\% in precision score, overtaking all previously published trackers with the same backbone. It is notable that training our DETRack is very easy:  only 60 trained epochs, which is only 10\% to 20\% of the previous SOTA trackers.

\textbf{LaSOT.}
LaSOT \cite{LaSOT} is a densely annotated large-scale dataset that contains 280 long-term video sequences for public evaluation.  We evaluate our DETRack on the test set to compare with previous SOTA trackers. From \cref{tab:main}, we find that our method acheives a comparable results, surpassing the OSTrack with the same backbone in all three matrics.  Specifically, DETRack achieves 69.0\% AUC score with merely 60 training epochs, which comsumes quite a few GPU hours compared with the other SOTA methods, demonstrateing the efficiency of our approach.


\textbf{NFS and UAV123.}
NFS~\cite{kiani2017need} comprises 100 video sequences featuring fast-moving objects, and is often used to test the robustness of tracking algorithms. UAV123~\cite{mueller2016benchmark}, with its 123 video sequences captured from a low-altitude unmanned aerial vehicle, poses challenges for long-term tracking due to its average sequence length of 915 frames. As reported in \cref{tab:addition}, our method consistently outperforms the baseline on both datasets. Moreover, our approach requires significantly fewer training epochs, leading to reduced GPU training time.



\textbf{Params, GFLOPs and Speed.}
We provide the GFLOPs with state-of-the-art trackers in \cref{tab:main}.  For more details about params, GFLOPs and speed, we compare our DETRack with the baseline OSTrack~\cite{ostrack} in \cref{tab:flops} on a Nvidia RTX 2080Ti GPU.  We re-implemented the OSTrack by replacing the MAE~\cite{mae} backbone with CAE~\cite{cae} for fair comparison.  No matter with or without saprisified technique like candidate elimination (CE)~\cite{ostrack},  our DETRack has a lower number of parameters and FLOPs than the tracker using the same backbone but with a convolutional head. Still, the implemention detials in engineering make our tracker run a bit slower.

\section{Ablation Study and Analysis}
We analyze the main properties of the DETRack framework. For the following experimental studies, we follow GOT-10k test protocol unless otherwise noted.


\subsection{Analysis on the Number of Decoder Layers.}

\begin{table}
    \centering
    \resizebox{0.9\linewidth}{!}{
    \begin{tabular}{c|ccc|c}
        \Xhline{2\arrayrulewidth}
        \multicolumn{1}{l|}{}                                                & \multicolumn{3}{c|}{AO(\%)} & \multirow{3}{*}{GFLOPs}          \\
        \cline{1-4}
        \multicolumn{1}{l|}{\diagbox{L$_\mathtt{test}$}{L$_\mathtt{train}$}} & 2                           & 3                       & 4    & \\
        \hline
        4                                                                    & -                           & -                       & 72.5 & 15.9\\
        3                                                                    & -                           & \textbf{72.9}           & 72.3 & 15.6\\
        2                                                                    & 71.4                        & 72.1                    & 71.7 & 15.3\\
        1                                                                    & 71.2                        & 71.4                    & 71.1 & 15.1\\
        \Xhline{2\arrayrulewidth}
    \end{tabular}}
    \caption{Results of the ablation study on the number of decoder layers. L$_\mathtt{train}$ represents the number of decoder layers under training. L$_\mathtt{test}$ means the number of first n layers are used during the inference.  The results of AO(\%) means the accuracy obtained by only using the first n layers of the decoder.  The experiment are reported on the GOT-10k test split following the one-shot protocol with 20 epochs.}
    \label{tab:decoder_layer}
\end{table}
We investigate the influence of varying numbers of decoder layers. As shown in \cref{tab:decoder_layer}, the decoder with 3 layers strikes the best trade-off between the efficiency and accuracy.  Decreasing the number of decoder layers hurts the performance significantly, \eg, a two-layer decoder decrease the accuracy of 1.5 AO\% to a three-layer decoder. Adding layers in the decoder, \eg, a four-layer decoder does not brings gains on performance but introducing redundant cumputation in terms of GFLOPs.

Benefits from applying auxiliary loss on each layer of decoder and sharing the prediction head among these layers, we can design a flexible configuration for activating different decoder layers between training and testing.  For instance, we can obtain the intermediate prediction in the early layer of the decoder as the final prediction, leaving the fianl layers un-performed.  For instance, only using first-two layers of a three-layer also achieves a acceptable performance as 72.1 AO\% as shown in ~\cref{tab:decoder_layer}.  Furthermore, we can only use the first layer in testing no matter how many layers we set for training.  The performance of all one-layer decoder configurations shown in ~\cref{tab:decoder_layer} surpass the convolutional baseline OSTrack(shown in the ~\cref{tab:main}).  The results demostrate that our tracker can adapt flexibly between efficiency and accuracy, even if the training is done.

\subsection{Denoising Strategy}

\begin{table}
    \centering
    \resizebox{\linewidth}{!}{

        \begin{tabular}{c|l|l|lll}
            \Xhline{2\arrayrulewidth}
            \multirow{2}{*}{\# }            & \multicolumn{1}{c|}{\multirow{2}{*}{Positive}} & \multicolumn{1}{c|}{\multirow{2}{*}{Negative}} & \multicolumn{3}{c}{GOT-10k}                                                                    \\
            \cline{4-6}
                                            & \multicolumn{1}{c|}{}                          & \multicolumn{1}{c|}{}                          & \multicolumn{1}{c}{AO}      & \multicolumn{1}{c}{SR$_{0.5}$} & \multicolumn{1}{c}{SR$_{0.75}$} \\
            \hline
            \textit{baseline}                        & -                                              & -                                              & 71.2                        & 80.1                           & 68.3                            \\
            {\textcircled{\footnotesize 1}} & embedding                                      & embedding                                      & 71.3                        & 80.3                           & 68.2                            \\
            {\textcircled{\footnotesize 2}} & center                                         & corner                                         & \textbf{72.9}               & \textbf{82.1}                  & \textbf{69.9}                   \\
            {\textcircled{\footnotesize 3}} & center                                         & outside                                        & 72.4                        & 81.9                           & \textbf{69.9}                   \\
            \Xhline{2\arrayrulewidth}
        \end{tabular}}
    \vspace{-2mm}
    \caption{Comparison of different label assignment for denoising part in the training. \textit{Baseline} indicates the tracker trained without any auxiliary denoising technique.  The column \textbf{Positve} and \textbf{Negative} represents the source feature for the positive and negative queries in the denoising part when applying denoising training.}
    \label{tab:denoising}
    \vspace{-2mm}
\end{table}
We also investigate the strategy for denoising training.
The origional denoising design for object detection (OD) adopts a embdeding layer to encode the positive and negative labels~\cite{dn-detr}.  The encoded classification label is static to the input image because the classes are usually fixed in OD.  However, in single object tracking (SOT), the class of the target is generic and dynamic.  As shown in ~\cref{tab:denoising}, the fixed embedding for classification label does not help improving the performance too much over the baseline which is trained without denoising, \eg, 71.4\% AO vs. 71.2\% AO in \textcircled{\footnotesize 1} and \textit{baseline}.  
The slight increase of performance is come from the localization part, \textit{i.e. }, the  bounding box denoising.  As shown in \textcircled{\footnotesize 2} of ~\cref{tab:denoising}, our proposed denoising strategy which the center feature is picked up as the positive query and the corner ones are selected as negative, improving the accuracy to 72.9\% AO, leading the baseline by 1.7 point.  The results also demostrate that dynamic selected features as denoising quiries are more suitable for SOT, compared with fixed and static label embeddings.  Compared with \textcircled{\footnotesize 3}, which randomly pick the feature pixels outside the GT bounding boxas as negative quiries, the corner features acting as hard negative samples closer to the GT boxes are more helpful to improve the performance, \eg, 72.9\% vs. 72.4\% AO.


\subsection{Label Assignment}

\begin{table}[t]
    \centering
    \resizebox{0.95\linewidth}{!}{
    \begin{tabular}{l|ccc|ccc} 
        \Xhline{2\arrayrulewidth}
        {\multirow{2}{*}{Method}} & \multicolumn{3}{c|}{GOT-10k} & \multicolumn{3}{c}{LaSOT} \\
        \cline{2-7}
        & AO$_\text{20}$ & AO$_\text{50}$ & AO$_\text{100}$ & AUC$_\text{60}$ & AUC$_\text{100}$ & AUC$_\text{300}$ \\
        \hline
        {\textcircled{\footnotesize 1}}center & 55.8 & 65.4 & 71.2 & 43.0 & 52.3 & 68.8 \\
        {\textcircled{\footnotesize 2}}Hungarian & 51.0 & 59.1 & 65.2 & 35.6 & 47.9 & 66.4 \\ 
        \cline{1-1}
        {\textcircled{\footnotesize 3}}hard label & 65.7 & 65.4 & 66.3 & 62.1 & 63.4 & 62.9 \\
        {\textcircled{\footnotesize 4}}loc. quality & 72.9 & 72.9 & 72.4 & 69.0 & 68.6 & 68.9 \\
        \Xhline{2\arrayrulewidth}
    \end{tabular}}
    \caption{Comparison with different label assignment (for un-denoising part). AO$_\text{k}$ and AUC$_\text{k}$ denotes for the performance obtained by training k epochs for the benchmark. }
    \label{tab:assignment}
    \vspace{-4mm}
\end{table}
Incorporating the decoder lets us set the training objective for classification, akin to convolution-based heads, such as centerness. From \cref{tab:assignment}, it's clear that one-to-one assignments, represented by centerness \textcircled{\footnotesize 1} and bipartite matching \textcircled{\footnotesize 2}, require a large number of training epochs to be effective, \eg, 55.8\% AO at 20 epoch to 71.2\% AO at 100 epoch in \textcircled{\footnotesize 1} on GOT-10k.  In contrast, one-to-many assignments (\textcircled{\footnotesize 3} and \textcircled{\footnotesize 4}) notably decrease the epochs required to reach peak performance. Using \textcircled{\footnotesize 3} to allocate each positive sample with a rigid label results in sub-optimal performance. However, emphasizing localization quality, as represented by IoU, considerably enhances both convergence rate and performance.  For instance, performance obtained in the early stage like 72.9\% AO at 20 epoch on GOT-10k and 69.0\% AUC at 60 epoch on LaSOT are higher than that with more epochs.


\subsection{Convergence}
\begin{figure}
  \centering
  \includegraphics[width=.95\linewidth]{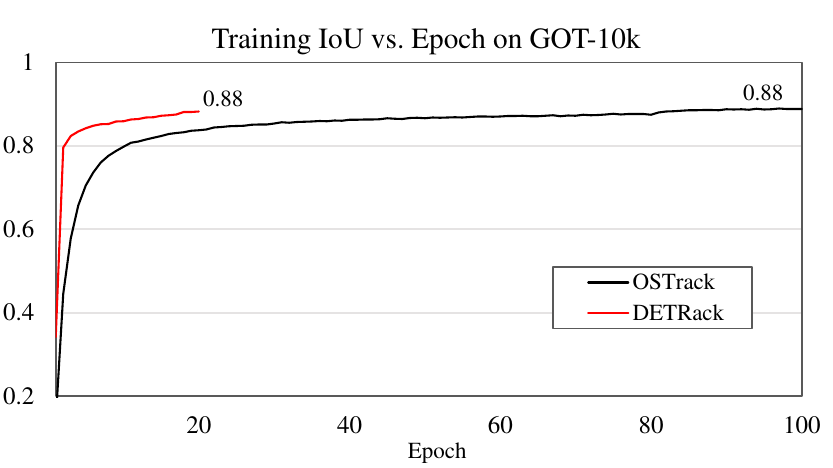}
  \vspace{-2mm}
  \caption{
    The label assignment for the denoising training.  The most center feature within the GT bounding box is picked as the positive object query in one denoising group, while the corner ones are selected as negative queries.
  }
  \label{fig:curve}
  \vspace{-3mm}
\end{figure}
We provide the training IoU vs. epoch curve in the \cref{fig:curve}.  All results are reported on 2 RTX 3090 GPUs with ViT-B intialized from CAE~\cite{cae}.  As we can see, our proposed DETRack achieve the similar training IoU in 20 epochs with the OSTrack in 100 epochs. The combined results from \cref{tab:assignment} and \cref{fig:curve} also demostrate a interesting fact: the more accurate localization ability in the template-search image pair during training does not mean a higher overall tracking performance over a video sequence in testing.

\section{Conclusion}
In this research, we unveiled DETRack, an innovative encoder-decoder framework that leverages the deformable transformer decoder to supersede the conventional convolutional head, paving the way for a more pronounced sparsity and consequent reduction in GFLOPs. Through our novel implementation of the one-to-many label assignment and unique denoising technique during training, DETRack improves tracking accuracy with a significantly reduced training epochs. The marked reductions in GPU hours for traiing are especially beneficial for researchers with limited computational resources, potentially democratizing access to high-quality visual object tracking.


\textbf{Limitation.}
Although our work achieves comparable accuracy on per-frame classification and localization with little GPU resource consuming, the overall performance on some long-term seqeunce tracking benchmark is limited.  Besides, even if our tracker runs with lower FLOPs and params than the convolution-based ones with the same or weaker performance, the actual running speed of ours, \eg, FPS, is a little lower than the latters due to the engineering implementation.  


{\small
\bibliographystyle{ieee_fullname}
\bibliography{egbib,grm_bib}
}

\end{document}